\documentclass[runningheads]{llncs}
\usepackage[T1]{fontenc}
\usepackage{graphicx}
\usepackage{hyperref}
\usepackage{makecell} 
\usepackage{wrapfig} 
\usepackage{amsmath,amssymb}
\usepackage{tikz}

\usepackage[inkscapelatex=false]{svg}
\usepackage{booktabs}
\usetikzlibrary{arrows.meta,positioning,shapes.geometric,fit,calc}

\newcommand{\todo}[1]{}
\renewcommand{\todo}[1]{{\color{red} {#1}}}

\begin{document}
\title{From Quotes to Concepts: Axial Coding of Political Debates with Ensemble LMs}
\titlerunning{From Quotes to Concepts}
% If the paper title is too long for the running head, you can set
% an abbreviated paper title here
\author{Angelina Parfenova\inst{1,2} \and
David Graus\inst{3} \and
Juergen Pfeffer\inst{1}}
\authorrunning{A. Parfenova et al.}
% First names are abbreviated in the running head.
% If there are more than two authors, 'et al.' is used.
%
\institute{Technical University of Munich \\
\email{angelina.parfenova@tum.de}\and
Lucerne University of Applied Sciences and Arts 
\\
 \and
University of Amsterdam\\
\email{d.p.graus@uva.nl}
}

\maketitle
\begin{abstract}
%reviewer 2 skepticism about FIRST METHOD
% We introduce the first method to perform \textit{axial coding} using large language models (LLMs), transforming raw debate transcripts into concise, hierarchical categories.
Axial coding is a commonly used qualitative analysis method that enhances document understanding by organizing sentence-level open codes into broader categories. In this paper, we operationalize \textit{axial} coding with large language models (LLMs). Extending an ensemble-based open coding approach with an LLM moderator, we add an axial coding step that groups open codes into higher-order categories, transforming raw debate transcripts into concise, hierarchical representations.
We compare two strategies: (i) clustering embeddings of code-utterance pairs using density-based and partitioning algorithms followed by LLM labeling, and (ii) direct LLM-based grouping of codes and utterances into categories.  We apply our method to Dutch parliamentary debates, converting lengthy transcripts into compact, hierarchically structured codes and categories. We evaluate our method using \textit{extrinsic} metrics aligned with human-assigned topic labels (ROUGE-L, cosine, BERTScore), and \textit{intrinsic} metrics describing code groups (coverage, brevity, coherence, novelty, JSD divergence). Our results reveal a trade-off: density-based clustering achieves high coverage and strong cluster alignment, while direct LLM grouping results in higher fine-grained alignment, but lower coverage ($\sim$20\%). Overall, clustering maximizes coverage and structural separation, whereas LLM grouping produces more concise, interpretable, and semantically aligned categories. To support future research, we publicly release the full dataset of utterances and codes, enabling reproducibility and comparative studies.

\end{abstract}


\section{Introduction}
% Introduce methods that support "SENSE-MAKING" of large textual corpora;
Automated methods for sense-making of large text corpora have a long tradition in information retrieval, in particular in domains where human review or exploratory search is central, such as legal search~\cite{cormack2010overview}, e-discovery~\cite{INR-025}, and patent retrieval~\cite{INR-027}.
%
% Introduces coding as established method in social sciences
In the social sciences, \emph{qualitative coding} has established itself as the de facto method for making large text collections interpretable. 
Here, human coders read text, assign short labels that capture its meaning (\emph{open coding}), and then compare, refine, and group these labels, typically across documents (\emph{axial coding}). 
While coding proves flexible in analyzing lengthy documents such as interview transcripts, political debates, or survey responses, it is known to be slow, subjective, and difficult to scale~\cite{Leeson2019}.

% Bridge to modern-day solutions: using LLMs
Recently, large language models (LLMs) have been shown to support qualitative coding by successfully generating plausible \emph{open codes} -- short descriptive labels for individual text segments~\cite{fischer-biemann-2024-exploring,parfenova-etal-2024-automating,tornberg2024large}. These approaches, however, omit the \emph{axial coding} step of grouping codes into higher-order categories, an abstraction stage that is central in qualitative analysis but has not been operationalized in prior IR or NLP methods.

% The abstraction step that human coders normally perform to organize codes into broader themes, known as axial coding, enables a broader overview and facilitates analysis.

\begin{figure}[t!]
\includegraphics[width=\columnwidth]{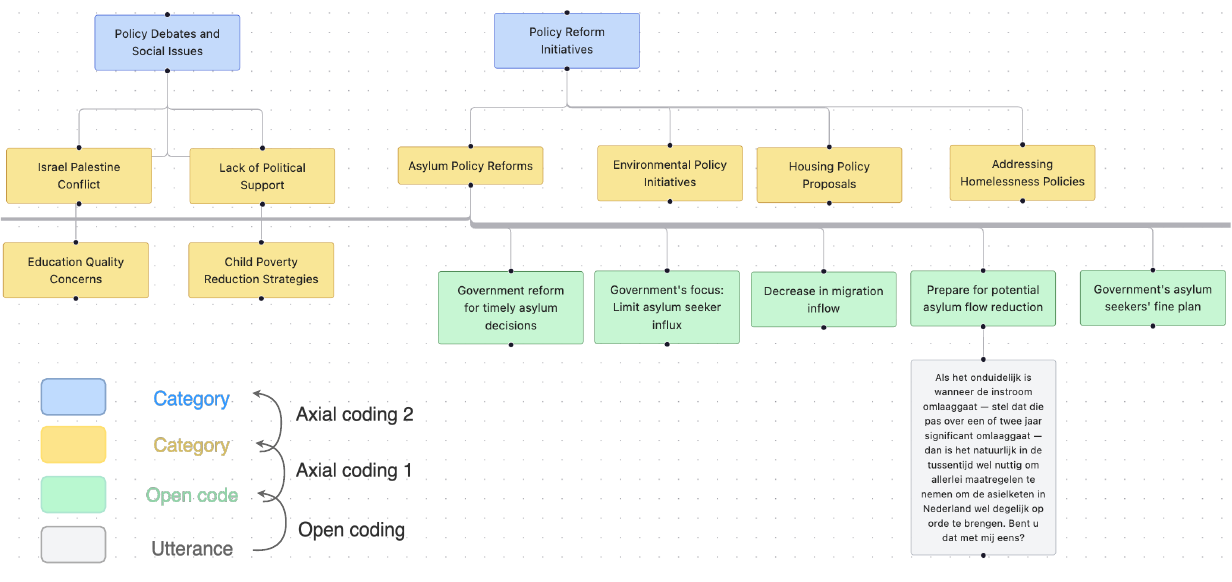}

    \caption{Example concept map based on the subset of results with HDBSCAN on the 5k held-out set. Utterances (grey) → open codes (green) → first-order
    axial categories (yellow) → second-order aggregates (blue).
    The second-order aggregates were obtained by applying the same grouping on the first-order categories; this is illustrative (not used in the evaluation), but helps convey the hierarchical nature of the axial coding process.}
    \label{fig:axial_coding_hierarchy}
\end{figure}

% Bring it home: show how we scale coding with LLMs, opening the door for large-scale coding of big corpora for human review
This paper introduces the first method that scales axial coding using LLMs, enabling the automated qualitative analysis of large-scale textual corpora. 
Our method extends an existing ensemble-based open coding framework~\cite{parfenova-pfeffer-2025-measuring} with an axial coding method to produce higher-order labeled code categories.
Hence, our method consists of two steps:
(i) \emph{open coding}: generate concise open codes using a moderator-style ensemble with label refinement, and 
(ii) \emph{axial coding}: group codes into higher-order labeled categories using one of two strategies: 
unsupervised clustering of embeddings (described in \S\ref{sec:clust-axial}), and LLM-based grouping (in \S\ref{sec:llm-axial}).
%
% \paragraph{Concept map (what the pipeline produces).}
To illustrate the output of our method, we show a compact concept map in Fig.~\ref{fig:axial_coding_hierarchy}, which visualizes how raw utterances (grey) are assigned open codes (green) and then grouped into axial categories (yellow). 
% These categories are the result of our two axial coding methods: clustering-based  and LLM-based (\S\ref{sec:llm-axial}), and are the focus of our evaluation.

\begin{figure}[t!]
\includegraphics[width=\columnwidth]{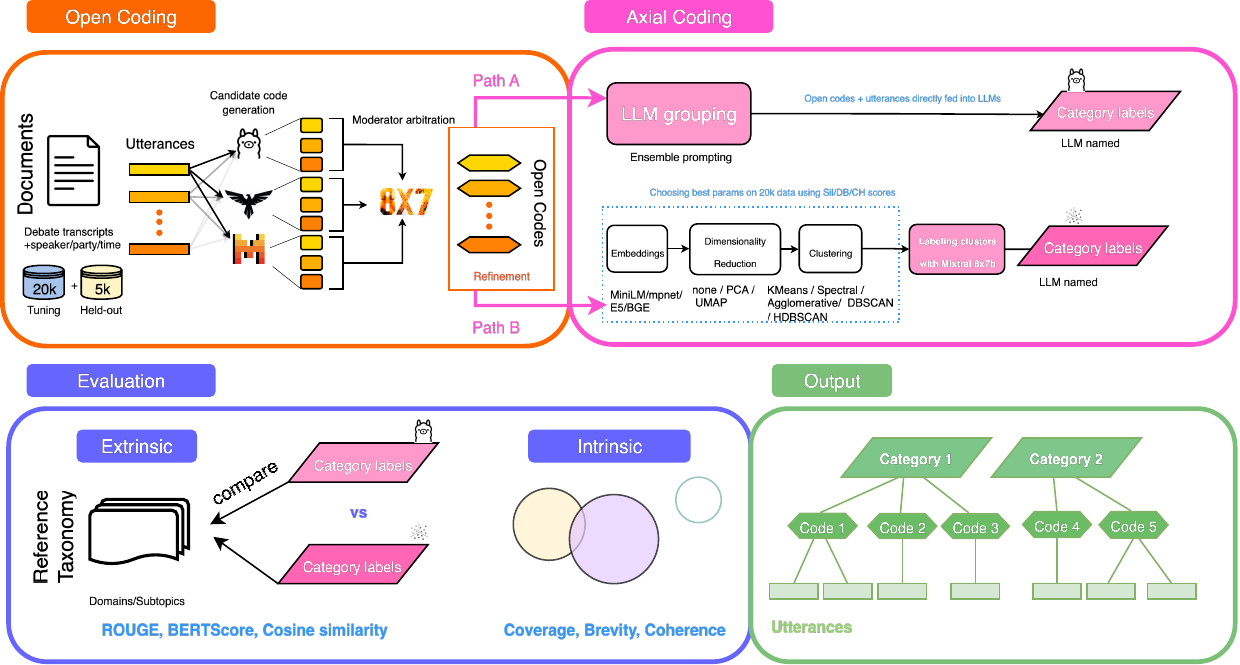}

    \caption{\textbf{Pipeline for open and axial coding.} 
In the open coding stage (orange box), we produce open codes for utterances from transcripts (20k train set for finding best parameters, 5k held-out set), using an ensemble of LoRA-finetuned LLMs with a moderator model, following the framework of \cite{parfenova-pfeffer-2025-measuring}. 
The axial coding stage (pink box) groups them into categories using one of two methods: direct grouping with LLM prompting (path A), or clustering embeddings, followed by LLM labeling (path B). 
Evaluation (blue box) covers both \emph{extrinsic} alignment with human-assigned domain labels and \emph{intrinsic} interpretability metrics (coverage, brevity, coherence, novelty, divergence). 
The output (green box) is a structured mapping from utterances to codes and categories, see also Fig.~\ref{fig:axial_coding_hierarchy}.}
    \label{fig:pipeline}
\end{figure}

We validate our method using transcripts of Dutch parliamentary debates, a challenging domain with heterogeneous speakers, procedural dialogue, evolving topics, and lengthy documents. 
To assess both the fidelity to human annotations and interpretability of the resulting groups, we evaluate our method using two approaches: 
\emph{extrinsic} evaluation, where we measure alignment with human-assigned labels using similarity metrics such as ROUGE \cite{lin-2004-rouge}, cosine similarity, and BERTScore \cite{bertscore}; 
and \emph{intrinsic} evaluation, where we use metrics independent of gold labels that capture properties deemed important for qualitative analysis, including coverage, label brevity, coherence, novelty, and divergence~\cite{chen2024computational}. 
This dual evaluation allows us to assess generated categories even where gold standards are partial or contested, highlighting how axial coding with LLMs can scale qualitative analysis while retaining interpretability.

% \todo{Our contribution is $n$-fold:
% (i) we re-implement an existing framework for LLM-based \emph{open coding}, and validate it on a novel domain (political debate transcripts) in a new language (Dutch),
% (ii) we extend the existing approach with an \emph{axial coding} stage, evaluating different methods for grouping \emph{open codes}, and validate these through gold standard evaluation and intrinsic evaluation, (iii) we publicly release our dataset of debate utterances, associated gold labels and LLM-generated open and axial codes, and
% ($n$) \ldots}

Our contributions are three-fold: (1) We introduce the first approach to scale \emph{axial coding} with LLMs, extending ensemble-based open coding with two alternative strategies: (i) clustering code embeddings with LLM-based labeling, and (ii) direct grouping of codes by LLMs.  (2) We design a dual evaluation scheme that combines \emph{extrinsic} alignment with human-coded taxonomies (ROUGE, cosine, BERTScore) and \emph{intrinsic} interpretability metrics (coverage, brevity, coherence, novelty, divergence), enabling systematic comparison of clustering- vs.\ LLM-based axial coding. (3) We publicly release a dataset of 5,000 debate utterances from Dutch parliamentary transcripts, including human-assigned domain/subtopic labels, LLM-generated open codes, and axial categories from both clustering and LLM methods, to support reproducibility and comparative studies.\footnote{\url{https://github.com/Likich/axial_coding_dataset}}

\section{Background}
Qualitative data analysis (QDA) is a central method in the social sciences to identify and interpret patterns in unstructured text~\cite{miller1990introduction,Creswell2016}. 
A core step is \emph{coding}: assigning short essence-capturing labels to text segments~\cite{Saldana2016}.
These codes form the building blocks for higher-level categories and themes~\cite{Braun2021}. 
Coding is typically an iterative and collaborative process, involving negotiation and refinement by teams of analysts. 
In QDA, two main stages of coding are commonly distinguished: 
(i) \emph{open coding} generates initial codes that capture the essence of text segments, and (ii) 
\emph{axial coding} organizes these codes into higher-order categories~\cite{Saldana2016}. Axial coding originates from grounded theory methodology and has been widely adopted in qualitative research for theory construction, category refinement, and comparative analysis across cases \cite{glaser2017discovery}. %Addressing reviewer 2
Coding can proceed \emph{inductively}, allowing codes to emerge organically from the material (in line with grounded theory traditions~\cite{glaser2017discovery}), or \emph{deductively}, by applying a predefined set of labels. 
Qualitative coding enables researchers to turn raw text into structured and interpretable data, and presents a complementary perspective to quantitative analysis methods commonly used in information retrieval~\cite{fidel1993qualitative}.

\paragraph{LLMs for qualitative coding.}
Recent work has explored large language models (LLMs) as tools to automate or support coding~\cite{Bommasani2021,laban-etal-2022-summac,parfenova-etal-2024-automating}. 
LLMs can produce plausible open codes and accelerate inductive or deductive coding \cite{fischer-biemann-2024-exploring,matter2024close,piano-etal-2023-qualitative,Xiao_2023,ziems2024can}. 
However, these methods have been limited to open coding, and no prior work has operationalized axial coding with LLMs to provide the abstraction step that groups codes to further facilitate analysis and human review.

\paragraph{Quantitative analysis of political discourse.}
The IR community has a long history in studying quantitative methods for analyzing and structuring large, heterogeneous text collections such as parliamentary debates. 
Foundational methods such as topic modeling \cite{blei2003latent} and embedding-clustering pipelines (e.g., BERTopic \cite{grootendorst2022bertopicneuraltopicmodeling}) offer unsupervised ways to identify topics and structures in large text corpora. 

% Addressing reviewer comments on Topic Modeling
While axial coding may appear superficially similar to coarse-grained topic modeling, the two differ in both intent and output. 
Topic models induce latent word distributions optimized for corpus-level structure \cite{blei2003latent,parfenova2024automating}, whereas axial coding operates over \emph{interpreted semantic units} (codes) and aims to construct human-readable, reusable conceptual categories. 
Axial categories are not latent variables but named abstractions intended for analytic reuse, comparison, and theory building. 
From this perspective, axial coding can be seen as an interpretive layer that operates on top of utterance-level codes rather than directly on token distributions, and thus complements topic modeling rather than replicating it.

Parliamentary debate graph extraction~\cite{salah_extracting_2013}, modeling \emph{“who said what to whom”}~\cite{kaptein_who_2009}, and visualization systems such as \textit{ThemeStreams}~\cite{de_rooij_themestreams_2013} and \textit{ATR-Vis}~\cite{makki_atr-vis_2018} demonstrate how IR methods can facilitate exploration and sense-making of political debates. 
Topic modeling-based methods are effective in identifying broad themes, but quality issues such as \emph{topic impurity} and \emph{topic generality} can make them difficult to interpret or reuse across studies~\cite{8482296}.
In addition, by focusing on coarse latent themes, quantitative methods risk losing the depth and contextual richness that qualitative methods can provide~\cite{fidel1993qualitative}. 
Our method induces fine-grained, interpretable codes at the utterance level, and organizes them into hierarchical categories, yielding structures that are more directly comparable to human-coded taxonomies, and potentially more useful for reuse and interpretive analysis.

\paragraph{Positioning.}
Our work builds on both the traditions of QDA in social sciences and the quantitative methods from IR: 
from QDA we take the notion of \emph{inductive} and \emph{axial coding}; from IR we inherit clustering methods, evaluation metrics, and graph-based modeling of debates. 
By combining ensemble LLM-based open coding with clustering- and LLM-driven axial coding, and by representing the results as a \emph{concept graph}, we propose a method that transforms raw debate transcripts into interpretable conceptual structures. 

\section{Dataset and Open Coding Method}
Our method extends an existing method for open coding using ensemble LLMs~\cite{parfenova-pfeffer-2025-measuring}, which we describe in~\S\ref{sec:open-coding}, below. 
We use a dataset created from Dutch parliamentary debate transcripts, with utterance-level metadata (speaker, party, role), document metadata (meeting title, date, \texttt{source\_doc\_id}), and associated topic labels from a human-curated taxonomy from the Dutch national government, consisting of 17 domains and 79 subtopics which are assigned at the document level.\footnote{\url{https://www.rijksoverheid.nl/onderwerpen/}}
We use these labels as ground truth for evaluation, rather than as direct supervision for coding. 
Table~\ref{tab:dataset-stats} summarizes key statistics of the datasets we use.

This corpus is particularly suitable for studying axial coding because debates are lengthy, multi-topic transcripts where abstraction into higher-order categories is naturally required for human review. In addition, the availability of a human-curated domain/subtopic taxonomy enables extrinsic evaluation without treating the taxonomy as supervision.
% Addressing Reviewer 3 

% This taxonomy provides our gold standard labels for evaluation. 

\begin{table}[h]
\centering
\scriptsize
\caption{Dataset statistics for the full corpus of Dutch parliamentary debates, the 20k training subset, and the 5k held-out test set. Number of utterances is reported per debate. Utterance length values are reported as mean / max words. 
Document length refers to the sum of all utterance lengths in a single debate. 
Speaker values are reported both globally and per debate. 
Topic counts are reported both at the domain level and the fine-grained subtopic level.}
\label{tab:dataset-stats}
\begin{tabular}{rrccrccclc}
\hline
\textbf{Split} & \textbf{Uttrs} & \textbf{Spkrs} & \textbf{Part.} & \textbf{Docs} & \makecell{\textbf{Spkrs}\\\textbf{per doc}\\(mean/max)} & \makecell{\textbf{Utt.}\\\textbf{per doc}} & \makecell{\textbf{Utt. len}\\(mean/max)} & \makecell{\textbf{Doc. len}\\(mean/max)} & \makecell{\textbf{Top./}\\\textbf{Subtop.}} \\
\hline
Full    & 82,732 & 248 & 17 & 2,038 & 4.7/77   & 40.6 & 119/7,060 & 4,844/124,352 & 17/79 \\
Train   & 20,000 & 211 & 17 &   578 & 4.0/59   & 34.6 & 138/7,060 & 4,760/83,863  & 17/53 \\
Test    &  5,000 & 128 & 15 &   196 & 3.8/57   & 25.5 & 125/2,022 & 3,178/50,129  & 16/42 \\
\hline
\end{tabular}
\end{table}

% \vspace{-10pt}

\subsection{Open Coding}
\label{sec:open-coding}
The first stage of our pipeline produces \textit{open codes} for each utterance: short labels that capture their essence. 
This first stage is illustrated in the orange box of Fig.~\ref{fig:pipeline}. 
We implement the ensemble-based coding framework of Parfenova and Pfeffer~\cite{parfenova-pfeffer-2025-measuring}, which mirrors how human coders typically work: multiple annotators propose candidate codes which are refined through team discussion to improve consistency. 
In our setup, multiple smaller specialized LLMs generate candidate codes, a moderator LLM arbitrates between them, and a post-hoc refinement step consolidates the outputs. 
This approach was shown to improve both stability and interpretability compared to using individual models alone. 

\paragraph{Candidate code generation.}
We use instruction-tuned LLMs (e.g., LLaMA-3, Falcon, Mistral), adapted to the open coding task using Low-Rank Adaptation (LoRA), with a dataset of 1,000 \textit{quote-code} pairs from social science research studies and the SemEval-2014 Task 4 dataset~\cite{pontiki-etal-2014-semeval}. 
Each model independently proposes a concise candidate label (1–5 words) for a given utterance, following prompts that encourage semantic abstraction rather than verbatim repetition.

\paragraph{Moderator arbitration.}  
The moderator is an LLM prompted to act as a lead coder: it is shown the original text segment and candidate codes from the fine-tuned open coding LLMs, and is asked to select the most appropriate one (or propose a refined alternative). 
In this way, the moderator provides a clear decision, and produces more consistent and stable code assignments than simple ensembling. 

\paragraph{Label refinement.}
To reduce code redundancy, we apply an iterative refinement process. 
Each newly generated code is compared against all previously assigned labels, using cosine similarity between SBERT embeddings. 
If the new code's similarity to a previous code exceeds a threshold ($\tau = 0.7$), the previous code is reused; otherwise, the new code is introduced. 
This threshold follows prior work on semantic similarity in qualitative coding, where 0.7 was empirically found to balance over-merging and under-merging labels~\cite{parfenova-pfeffer-2025-measuring}. 
This mechanism promotes label reuse, resulting in shorter average code lengths and more consistent codes.

\paragraph{Outputs.}
The result of this process is a single refined open code per utterance.

\section{Axial Coding Method}
The open codes resulting from the coding stage described above serve as input to our novel axial coding stage, illustrated in the pink box in Fig.~\ref{fig:pipeline}. 
We compare two methods for grouping codes into higher-order categories: (i) \emph{unsupervised clustering} in embedding space, and (ii) \emph{LLM-powered grouping}, where LLMs directly propose and name categories. 

\subsection{Clustering-based axial coding}\label{sec:clust-axial}
We implement a clustering-based axial coding method, by embedding both the \textit{open code} and its associated \textit{utterance} jointly. 
This balances interpretability of concise labels with richer contextual information.

For embedding the input, we tested a range of sentence encoders: \texttt{all-MiniLM-\\L6-v2}, \texttt{multi-qa-MiniLM-L6-cos-v1}, \texttt{paraphrase-MiniLM-L3-v2}, \texttt{all-mpnet-\\base-v2}, and the \texttt{e5} and \texttt{bge} families (small/base/large). 
Dimensionality reduction was varied between none, PCA (64 dimensions), and UMAP (15 dimensions). 
For the actual clustering, we compared algorithms including KMeans, Agglomerative, Spectral, Gaussian Mixture Models (GMM), and density-based methods: DBSCAN and HDBSCAN. 
For density-based methods, we use \textit{ms} to denote DBSCAN’s \texttt{min\_samples} parameter, and \textit{mcs} to denote HDBSCAN’s \texttt{min\_cluster\_size}. 
For all other methods, we fix the number of clusters to $k \in \{4,6,\dots,20\}$, which roughly aligns with the 17 domain-level categories in the human taxonomy. 
This mirrors topic modeling workflows while retaining code-level interpretability.

To identify which clustering approach performs best, we first conduct a large-scale sweep on the 20k training set, ranking by intrinsic clustering metrics: silhouette score, Davies–Bouldin Index (DBI), and Calinski–Harabasz Index (CHI). 
Top-performing configurations are reported in Table~\ref{tab:train-top-configs}, and reapplied to the 5k held-out set to assess whether they generalize well to unseen utterances (Table~\ref{tab:test5k-clusters}).

\paragraph{Results.}
We find that density-based methods (DBSCAN and HDBSCAN) achieve the highest silhouette scores, reflecting compact, well-separated clusters.
However, this comes at the cost of coverage: DBSCAN often discards over 85–95\% of utterances as noise. 
The resulting clusters are small and clean, but unrepresentative of the debate corpus. 
Other clustering models do not discard utterances as much, providing full coverage, but yield less separated clusters according to the Silhouette score.  
Overall, we adopt both density-based clustering methods (DBSCAN and HDBSCAN) in our axial coding method, since they balance coverage with flexible discovery of cluster counts, while still achieving strong internal scores. 
The high noise rates of density-based methods reflect conservative cluster formation rather than failure, mirroring qualitative coding practices where ambiguous material may be left uncategorized during early abstraction stages. %Addressing “Density-based clustering discards most utterances.”

\begin{table}[h]
\centering
\caption{Top clustering configurations on the 20k training set. 
For KMeans, Agglomerative, Spectral, and GMM we report the standard silhouette score. 
For density-based methods (DBSCAN, HDBSCAN), we report silhouette computed only on core points (excluding noise), since global silhouette is dominated by the noise label.}
\label{tab:train-top-configs}
\resizebox{\textwidth}{!}{%
\begin{tabular}{llllclrrr}
\hline
\textbf{Embedding} & \textbf{Reduction} & \textbf{Algo} & \textbf{Params} &
\textbf{Silhouette} & \textbf{DBI} & \textbf{CHI} & \textbf{Clusters} & \textbf{Noise rate} \\
\hline
all-MiniLM-L6-v2        & none & DBSCAN        & eps=0.3, min\_samples=20 & 0.747 & 1.057 &   65.6  & 18  & 0.956 \\
all-mpnet-base-v2       & none & DBSCAN        & eps=0.3, min\_samples=20 & 0.731 & 1.097 &   65.8  & 17  & 0.955 \\
all-MiniLM-L6-v2        & none & DBSCAN        & eps=0.3, min\_samples=10 & 0.707 & 1.066 &   40.2  & 41  & 0.939 \\
paraphrase-MiniLM-L3-v2 & UMAP & HDBSCAN       & in\_cluster\_size=10    & 0.668 & 0.786 &  278.0  & 302 & 0.504 \\
all-MiniLM-L6-v2        & UMAP & HDBSCAN       & min\_cluster\_size=20    & 0.667 & 1.026 &  265.0  & 151 & 0.507 \\
bge-base-en             & UMAP & HDBSCAN       & min\_cluster\_size=50    & 0.664 & 1.086 &  514.2  &  55 & 0.560 \\
e5-small                & UMAP & KMeans        & k=4, n\_clusters=4            & 0.571 & 0.615 & 11982   &  4  & 0     \\
e5-small                & UMAP & GMM           & k=4, n\_components=4          & 0.530 & 0.687 & 9213    &  4  & 0     \\
all-mpnet-base-v2       & UMAP & Spectral      & k=4, n\_clusters=4, affinity=nn & 0.484 & 1.476 &  112.6 &  4  & 0     \\
e5-small                & UMAP & Agglomerative & k=8, n\_clusters=8            & 0.477 & 0.701 & 9957    &  8  & 0     \\
\hline
\end{tabular}
}
\end{table}

\begin{table}[t]
\centering
\caption{Performance of top clustering configurations on the 5k held-out set. 
Silhouette is computed only on core points (excluding noise), since global silhouette is dominated by the noise label. 
Other values include Davies–Bouldin Index (DBI), Calinski–Harabasz Index (CHI), number of clusters, number of noise points, and proportion of points labeled as noise.}
\label{tab:test5k-clusters}
\resizebox{\textwidth}{!}{%
\begin{tabular}{llccrrrr}
\hline
\textbf{Algorithm} & \textbf{Params} & \textbf{Silhouette} & \textbf{DBI} & \textbf{CHI} & \textbf{Clusters} & \textbf{Noise points} & \textbf{Noise rate} \\
\hline
DBSCAN  & eps=0.3, min\_samples=20  & 0.809 & 0.979 & 111.6 & 6  & 4,484 & 0.897 \\
DBSCAN  & eps=0.3, min\_samples=10  & 0.657 & 1.037 &  63.5 & 13 & 4,390 & 0.878 \\
HDBSCAN & min\_cluster\_size=20     & 0.646 & 1.119 & 310.9 & 49 & 2,394 & 0.479 \\
DBSCAN  & eps=0.3, min\_samples=20  & 0.635 & 1.079 &  93.1 & 7  & 4,433 & 0.887 \\
HDBSCAN & min\_cluster\_size=10     & 0.612 & 0.905 & 488.3 & 83 & 1,985 & 0.397 \\
HDBSCAN & min\_cluster\_size=50     & 0.511 & 1.283 & 1417.2 & 3 & 200  & 0.040 \\
\hline
\end{tabular}}
\end{table}

\subsection{LLM-based axial coding}\label{sec:llm-axial}

Each utterance–code pair is formatted as a compact item string: 
\mbox{\texttt{Code: "…"}} \mbox{\texttt{Utterance: "…"}}. 
We process the corpus in medium-sized batches (200–500 items) to balance context window limits with coverage, and pass each batch to an LLM with the instruction to \emph{group items into higher-level categories}. 
The model returns a JSON object of the form
\texttt{\{"category\_label": [code\_1, code\_2, …], …\}}, with category labels limited to at most five words.  

We run three LLMs independently over the full corpus: Deepseek-R1, Llama-4 Maverick, and Mixtral 8$\times$7B. 
The outputs across batches are then merged using 
\emph{label similarity} (cosine similarity between SBERT embeddings of category labels, using \textit{all-MiniLM-L6-v2}) and 
\emph{membership overlap} (Jaccard similarity between item sets).

Categories are merged if their labels reach a similarity threshold ($\geq 0.70$ cosine, similarly to open code merging) or a memberships overlap threshold ($\geq 0.20$ Jaccard). 
From each merge set, the shortest category label is chosen as the representative name. 
% The resulting categories are assigned stable identifiers. 
Each utterance is placed into a single category, defaulting to the largest category in cases of multiple assignments. Importantly, LLMs are not forced to assign each utterance to a category. In practice, some utterances may be left ungrouped when the model does not include them in any category, or when they are not retained during the category merging step. 
These are marked as \emph{Unassigned} and counted in our \textbf{Coverage} metric, reflecting the expected “abstention” behavior of LLMs. 
This abstention behavior mirrors human axial coding practice, where not all codes are immediately assigned to higher-order categories, particularly when semantic fit is weak or ambiguous.

\section{Evaluation}
We assess axial coding performance along two complementary dimensions. 
Extrinsic evaluation measures alignment with the human-assigned labels of our dataset, providing a reference-based benchmark. 
Intrinsic evaluation operates without reference labels and relies on unsupervised metrics that capture properties of the extracted categories that are central to qualitative data analysis.

%Motivation for choosing these metrics more

\paragraph{Intrinsic evaluation.}  
First, we evaluate categories independently of a gold taxonomy, focusing on properties that matter for qualitative coding \cite{chen2024computational}. 
We use the following measures:
\begin{itemize}
    \item \textbf{Coverage}: the proportion of utterances assigned to categories.
    \item \textbf{Brevity}: the average number of words per category label. 
    \item \textbf{Coherence}: the mean textual similarity of utterances within the same category, estimated from MiniLM embeddings.
    \item \textbf{Novelty}: the proportion of singleton categories.
    \item \textbf{Divergence}: the difference of the cluster-size profile (i.e., distributions) of one method with the reference distribution of aggregated cluster sizes. 
\end{itemize}

We measure \textbf{divergence} by comparing each method’s category-size distribution to the \emph{aggregated code space} (ACS), defined as the mean normalized size distribution across all methods (taken from~\cite{chen2024computational}). 
Divergence from this distribution is quantified using the Jensen–Shannon distance (JSD), a symmetric measure of distributional difference. 
A lower JSD indicates that method’s categories are more stable and aligned with the ensemble norm.

Together, these metrics operationalize core qualitative concerns: representativeness (coverage), interpretability (brevity), semantic consistency (coherence), abstraction balance (novelty), and structural stability (divergence), enabling evaluation of category systems without relying on a fixed gold taxonomy.

% To compute \textbf{divergence}, we define the \emph{aggregated code space} (ACS) as the mean normalized cluster-size distribution across all methods. 
% If $s^{(m)} \in \mathbb{R}^L$ is the normalized size vector for axial coding method $m$, sorted and padded to length $L$, then
% \[
% ACS = \frac{1}{M} \sum_{m=1}^M s^{(m)}
% \]
% denotes the ACS. 
% For a given method with distribution $p$, stability is measured via the Jensen–Shannon distance
% \[
% \text{JSD}(p \| ACS) = \tfrac{1}{2} D_{\mathrm{KL}}\!\left(p \;\middle\|\; \tfrac{p+ACS}{2}\right) + \tfrac{1}{2} D_{\mathrm{KL}}\!\left(ACS \;\middle\|\; \tfrac{p+ACS}{2}\right),
% \]
% where $D_{\mathrm{KL}}$ is the Kullback–Leibler divergence. 
% Lower JSD indicates closer alignment with the ensemble norm, reflecting more stable and generalizable category structures. 

\paragraph{Extrinsic evaluation.}  
Second, we compare our method's outputs to the human-coded taxonomy of debate topics, which consists of two levels: \emph{domains} (16 in our test set) and \emph{subtopics} (42).
We compute ROUGE (1/2/L), cosine similarity using MiniLM embeddings, and BERTScore~F1, reporting results at both the domain and subtopic level. 
Importantly, this evaluation relies on (textual) similarity metrics, and not, e.g., classification metrics or accuracy, as the labels produced by our method are inductive codes generated by LLMs, not pre-existing (sub)selected from the taxonomy. 

%\vspace{-10pt}

\section{Results}
% \vspace{-3pt}

\paragraph{Intrinsic evaluation}  
Intrinsic metrics in Table~\ref{tab:intrinsic_eval} reveal a coverage–coherence trade-off. 
DBSCAN delivers high intra-cluster coherence (0.93--0.95) but at extremely low coverage (0.10--0.12), making its categories unrepresentative of the corpus. 
HDBSCAN increases coverage substantially: the BGE+UMAP configuration achieves 96\% coverage and the lowest divergence (0.526), but collapses the data into a mere three categories, with moderate coherence (0.661). 
More balanced settings (e.g., MiniLM+UMAP, Paraphrase+UMAP) discover 49--80 categories with coherence around 0.68--0.69 and similarly moderate divergence, providing finer partitions while avoiding extreme overmerging. 

For LLM grouping (bottom three rows), labels are comparatively more concise (1.8--2.9 words on average), and categories align relatively well with consensus size distributions (ACS). 
Deepseek-R1 shows the lowest divergence (0.134), meaning its partitioning is structurally most similar to the consensus. 
Llama-4 achieves the highest weighted coherence of the LLM methods (0.576) and the overall lowest novelty (0.024), suggesting cleaner, less fragmented categories, though its coverage (0.328) remains far below the best clustering methods.

\begin{wrapfigure}{r}{0.6\textwidth}  % r=right, l=left
                        % pull up to align with heading (tweak)
  \centering
  %\vspace{-45pt}
  \includegraphics[width=\linewidth]{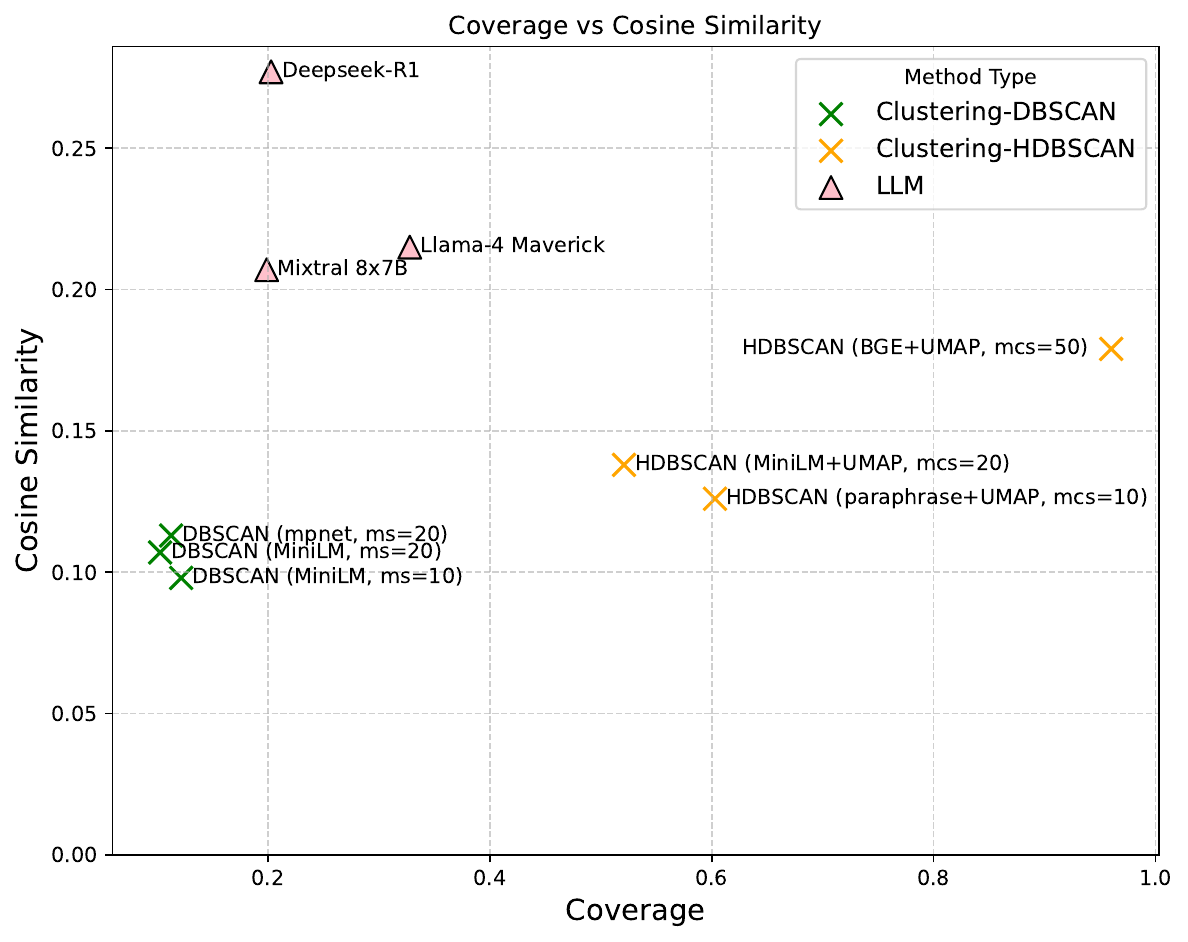} % your file name
  \caption{Coverage vs.\ cosine similarity for clustering- and LLM-based axial coding methods. 
    Pink triangles denote LLM-based grouping, green crosses denote DBSCAN variants, 
    and orange crosses HDBSCAN variants.}
    \vspace{-10pt}
  \label{fig:covcos}
  \vspace{-2pt}                       % tighten space to following text
\end{wrapfigure}

\paragraph{Extrinsic evaluation}  
Next, we compare model outputs with the human-labeled domains and subtopics from the taxonomy, see Table~\ref{tab:extrinsic_eval}. 
Among clustering methods, HDBSCAN with BGE embeddings and UMAP (\(mcs{=}50\)) achieves the strongest alignment on the held-out set.

At the subtopic level, it reaches ROUGE-L of 0.111, cosine similarity of 0.300, and BERTScore of 0.867, while covering 96\% of utterances (4,800 of 5,000). 
In contrast, DBSCAN variants achieve near-zero ROUGE because they label around 90\% of utterances as noise.  

LLM-based grouping improves fine-grained alignment further: Deepseek-R1 attains a ROUGE-L score of 0.248 and cosine similarity of 0.328 at the subtopic level, outperforming the best clustering run on both metrics. 
However, it assigns categories to only about 20\% of utterances, limiting coverage. 
Llama-4 Maverick and Mixtral 8$\times$7B achieve lower ROUGE-L overall, though BERTScore remains uniformly high and less discriminative across models.

\paragraph{Summary.}  
Overall, clustering methods, in particular HDBSCAN, offer strong coverage and control over granularity but risk summarizing debates into overly broad categories. 
LLM grouping, on the contrary, produces compact and interpretable labels and achieves the strongest ROUGE scores but leaves a large portion of utterances unassigned.
This trade-off between coverage and external alignment is illustrated in Fig.~\ref{fig:covcos}, where we plot coverage on the $x$-axis and similarity on the $y$-axis.
% \todo{The figure shows how DBSCAN attains relatively high coherence but discards most utterances (resulting in low Coverage), HDBSCAN balances coverage and similarity with some over-merging (?), while LLMs achieve competitive alignment with interpretable labels (similarity) but cover fewer utterances (low coverage).}
The figure highlights a clear coverage–alignment trade-off between clustering-based and LLM-based axial coding methods.
HDBSCAN variants (orange crosses) achieve high coverage, with moderate alignment to the gold taxonomy.
LLM-based methods (pink triangles) show the opposite pattern: substantially higher similarity, at the cost of much lower coverage.
DBSCAN variants (green) perform comparatively poorly on both dimensions, exhibiting especially low coverage by discarding most utterances.

\paragraph{Data availability.}
We release the 5k test subset coded by the best clustering configurations and LLMs, including utterance text, human domain/subtopic labels, open codes, and axial categories.\footnote{\url{https://github.com/Likich/axial_coding_dataset}}

\begin{table*}[t]
\centering
\caption{Intrinsic evaluation results, showing Coverage (Cov), number of categories (\#Cats), Brevity (Brev), Coherence (Coh), Novelty (Nov), and Divergence (Div) for both clustering and LLM-based axial coding methods. Best values across all methods are shown in bold.}
\label{tab:intrinsic_eval}
\resizebox{\textwidth}{!}{%
\begin{tabular}{l l r r r r r r}
\toprule
\textbf{Method (params)} & \textbf{Type} 
& \textbf{Cov} & \textbf{\#Cats} & \textbf{Brev} & \textbf{Coh} & \textbf{Nov} & \textbf{Div} \\
\midrule
DBSCAN (MiniLM, ms=10)            & Clust. & 0.122 & 13 & 3.25 & 0.932 & 0.385 & 0.712 \\
DBSCAN (MiniLM, ms=20)            & Clust. & 0.103 &  6 & 3.29 & \textbf{0.953} & 0.500 & 0.708 \\
DBSCAN (mpnet,  ms=20)            & Clust. & 0.113 &  7 & 3.00 & 0.939 & 0.143 & 0.700 \\
HDBSCAN (paraphrase+UMAP, mcs=10) & Clust. & 0.603 & 80 & 3.27 & 0.690 & 0.588 & 0.567 \\
HDBSCAN (MiniLM+UMAP, mcs=20)     & Clust. & 0.521 & 49 & 3.26 & 0.676 & 0.490 & 0.575 \\
HDBSCAN (BGE+UMAP, mcs=50)        & Clust. & \textbf{0.960} &  3 & 3.05 & 0.661 & 0.333 & 0.526 \\
\midrule
Deepseek-R1                        & LLM    & 0.203 & 67 & 2.90 & 0.478 & 0.239 & \textbf{0.134} \\
Llama-4 Maverick                   & LLM    & 0.328 & 83 & 2.30 & 0.576 & \textbf{0.024} & 0.225 \\
Mixtral 8$\times$7B                & LLM    & 0.199 & 33 & \textbf{1.80} & 0.499 & 0.121 & 0.260 \\
\bottomrule
\end{tabular}}
\end{table*}

\begin{table*}[t]
\centering
\caption{Extrinsic evaluation results, showing ROUGE-L (RL), Cosine similarity (Cos), and BERTScore (BERT) at the Domain and Subtopic level for both clustering and LLM-based axial coding methods. Best values across all methods are in bold.}
\label{tab:extrinsic_eval}
\resizebox{\textwidth}{!}{%
\begin{tabular}{l l r r r r r r}
\toprule
\textbf{Method (params)} & \textbf{Type} 
& \multicolumn{3}{c}{\textbf{Domain}} 
& \multicolumn{3}{c}{\textbf{Subtopic}} \\
\cmidrule(lr){3-5} \cmidrule(lr){6-8}
& & RL & Cos & BERT & RL & Cos & BERT \\
\midrule
DBSCAN (MiniLM, ms=10)            & Clust. & 0     & 0.098 & 0.832 & 0.008 & 0.121 & 0.835 \\
DBSCAN (MiniLM, ms=20)            & Clust. & 0     & 0.107 & 0.828 & 0     & 0.143 & 0.832 \\
DBSCAN (mpnet,  ms=20)            & Clust. & 0     & 0.113 & 0.840 & 0     & 0.148 & 0.844 \\
HDBSCAN (paraphrase+UMAP, mcs=10) & Clust. & 0.004 & 0.126 & 0.835 & 0.028 & 0.169 & 0.842 \\
HDBSCAN (MiniLM+UMAP, mcs=20)     & Clust. & 0.015 & 0.138 & 0.835 & 0.038 & 0.183 & 0.842 \\
HDBSCAN (BGE+UMAP, mcs=50)        & Clust. & 0     & 0.179 & 0.846 & 0.111 & 0.300 & 0.867 \\
\midrule
Deepseek-R1                        & LLM    & \textbf{0.076} & \textbf{0.277} & 0.853 & \textbf{0.248} & \textbf{0.328} & 0.859 \\
Llama-4 Maverick                   & LLM    & 0.015 & 0.215 & 0.853 & 0.057 & 0.252 & 0.862 \\
Mixtral 8$\times$7B                & LLM    & 0.004 & 0.207 & \textbf{0.872} & 0.001 & 0.187 & \textbf{0.871} \\
\bottomrule
\end{tabular}}
\end{table*}

\section{Discussion}

The two axial coding strategies show complementary strengths. 
Clustering, particularly HDBSCAN with BGE embeddings and UMAP, achieve near-complete coverage and moderate external alignment, but often at the cost of overly broad categories (for example, only three clusters in the best configuration). 
DBSCAN variants discard most of the utterances as noise and thus fail to align with the taxonomy.  

LLM-based grouping, on the contrary, produces compact and interpretable categories, with succinct labels (1.8--2.9 words), and lower divergence from the aggregated code space. 
The best-performing LLM (Deepseek-R1) outperforms the clustering methods in external alignment with subtopics, though with much lower coverage. 
Together, clustering methods optimize \emph{coverage and structural separation}, while LLMs excel in \emph{label quality and semantic alignment}. 
A hybrid approach using clustering for structural grouping and LLMs for semantic summarization offers the most balanced trade-off for axial coding of debates. 
A concrete pipeline would proceed in two stages: (i) apply clustering to ensure full corpus coverage and establish coarse structural partitions, and (ii) apply LLM-based summarization within each cluster to generate concise, interpretable axial labels. 
This preserves recall while leveraging LLMs where they are strongest—semantic abstraction and naming. 
We leave empirical validation of this hybrid strategy to future work. % Addressing reviewer question on exact hybrid approach

Our results suggest that different axial coding methods may optimize distinct retrieval scenarios and information needs. 
Clustering methods, with broad coverage and structural separation, provide a strong basis for recall-oriented search, where the goal is to capture all potentially relevant utterances under coarse themes. 
LLM-based grouping, with concise and interpretable labels, may better support precision-oriented search, surfacing categories that are meaningful and user-friendly for downstream search or recommendation tasks. 
Thus, a hybrid pipeline could bridge IR needs by combining clustering for exhaustive indexing with LLM summarization for interpretability.

%Seems contradictory, rewrite more understandable 

\subsection{Limitations and future work}
Our study is limited in several respects. 
First, we focus on a single debate corpus, which constrains the generalizability of findings. 
Political debates have distinctive structures and rhetorical styles, and it remains to be tested whether our methods extend to other qualitative domains such as interviews, focus groups, or ethnographic texts. 
Second, we evaluate only a fixed set of LLMs and clustering algorithms; more recent or larger models, alternative prompting strategies, and other clustering approaches could yield different outcomes. 
Third, while our evaluation covers both extrinsic and intrinsic metrics, it still assumes a relatively clean gold taxonomy; future work should address settings where taxonomies are partial, evolving, or contested. 
Finally, although we illustrate the potential of hierarchical concept graphs in Fig.~\ref{fig:axial_coding_hierarchy}, our experiments emphasize only the first stage of axial coding. 
Deeper hierarchies, dynamic graph construction, and applications such as temporal theme tracking, cross-party comparisons, or integration with political information retrieval systems remain to be explored.

\section{Conclusion}

This paper introduces a method for transforming raw utterances in political debates into structured conceptual representations via open and axial coding with ensemble LLMs. 
By combining a moderator-based open coding stage with clustering-based grouping and direct LLM-based grouping, we demonstrate how inductive coding can be scaled while retaining interpretability. 

Our results reveal complementary strengths. 
Density-based clustering methods discover cluster counts close to the annotated domains and yield high internal separation, but exclude a substantial portion of utterances as noise in comparison to partition-based methods.  However, even if partition-based clustering methods ensure full coverage of utterances, they require fixing the number of clusters in advance, and tend to sacrifice interpretive quality. 
Direct LLM grouping, on the other hand, while achieving lower coverage than HDBSCAN, adapts flexibly to the data and produces concise, human-readable labels that better support interpretation. 

Overall, our experiments show that clustering provides broad structural coverage of debates, while LLM grouping produces compact and semantically meaningful categories. 
Rather than treating these as competing strategies, we view them as complementary: clustering ensures that a vast number of utterances are represented, and LLMs supply the interpretive layer that makes categories comprehensible.

\subsubsection*{Acknowledgements.}
David Graus is partly funded by ICAI (AI for Open Government Lab). Views expressed in this paper are not necessarily shared or endorsed by those funding the research.

\subsubsection*{Disclosure of Interests}
The authors have no competing interests to declare that are relevant to the content of this article.

\bibliographystyle{splncs04}
\bibliography{custom}

\end{document}